\documentclass[sigconf]{acmart}

\usepackage{multirow}
\definecolor{electricviolet}{rgb}{0.56, 0.0, 1.0}
\usepackage{amsmath}
\usepackage{amsfonts}

\usepackage{subcaption}

\usepackage{url}
\usepackage{textcomp}
\definecolor{prismgreen}{rgb}{0, 0.6, 0}
\newcommand{\prismfont}{\fontsize{6pt}{6.9pt}\selectfont}

\usepackage{graphicx}
\usepackage{xcolor}
\usepackage{listings}

\lstdefinelanguage{Prism}{ 
basicstyle=\color{black}\prismfont\sffamily, 
keywords={bool,C,ceil,const,ctmc,double,dtmc,endinit,endmodule,endrewards,endsystem,F,false,floor,formula,G,global,I,init,int,label,max,mdp,pomdp,min,module,nondeterministic,Pmin,Pmax,prob,probabilistic,R,rate,rewards,Rmin,Rmax,S,stochastic,system,true,U,X,observables,endobservables},
keywordstyle={\bfseries\color{black}},
numberstyle=\tiny\color{black},
comment=[l] {//}, morecomment=[s]{/*}{*/}, 
commentstyle= \color{prismgreen}, 
tabsize=4, 
captionpos=b, 
escapechar=@, 
literate=%
        {-}{{\textcolor{black}{$-$}}}{1}%
        {->}{{\textcolor{black}{$\rightarrow{}$}}}{2}%
        {0}{{\textcolor{blue}{0}}}{1}%
             {1}{{\textcolor{blue}{1}}}{1}%
             {p1}{{\textcolor{red}{p1}}}{1}%
             {2}{{\textcolor{blue}{2}}}{1}%
             {p2}{{\textcolor{red}{p2}}}{1}%
             {3}{{\textcolor{blue}{3}}}{1}%
             {4}{{\textcolor{blue}{4}}}{1}%
             {5}{{\textcolor{blue}{5}}}{1}%
             {6}{{\textcolor{blue}{6}}}{1}%
             {7}{{\textcolor{blue}{7}}}{1}%
             {8}{{\textcolor{blue}{8}}}{1}%
             {9}{{\textcolor{blue}{9}}}{1}%
             {.}{{\textcolor{blue}{.}}}{1}%
             {=}{{\textcolor{black}{=}}}{1}%
             {[}{{\textcolor{black}{[}}}{1}%
             {]}{{\textcolor{black}{]}}}{1}%
             {;}{{\textcolor{black}{;}}}{1}%
             {+}{{\textcolor{black}{+}}}{1}%
             {*}{{\textcolor{black}{*}}}{1}%
             {:}{{\textcolor{black}{:}}}{1}%
             {\&}{{\textcolor{black}{\&}}}{1}%
             {|}{{\textcolor{black}{|}}}{1}%
             {?}{{\textcolor{black}{?}}}{1}%
             {"}{{\textcolor{black}{"}}}{1}%
             {(}{{\textcolor{black}{(}}}{1}%
             {)}{{\textcolor{black}{)}}}{1}%
             {'}{{\textcolor{black}{'}}}{1}%
}

\newcommand{\change}[2]{%
  \textcolor{black}{#2}%
}

\newcommand{\approach}{COMPASS}

\newcommand{\chBlue}[1]{\textcolor{black}{#1}}

\AtBeginDocument{
  }

\copyrightyear{2026}
\acmYear{2026}
\setcopyright{cc}
\setcctype{by}
\acmConference[SEAMS '26]{21st International Conference on Software Engineering for Adaptive and Self-Managing Systems}{April 13--14, 2026}{Rio de Janeiro, Brazil}
\acmBooktitle{21st International Conference on Software Engineering for Adaptive and Self-Managing Systems (SEAMS '26), April 13--14, 2026, Rio de Janeiro, Brazil}
\acmDOI{10.1145/3788550.3794879}
\acmISBN{979-8-4007-2445-9/2026/04}

\begin{document}

\title{Mind the Prompt: Self-adaptive Generation of Task Plan Explanations via LLMs}

\author{Gricel V\'{a}zquez}
\orcid{0000-0003-4886-5567}
\affiliation{%
  \institution{University of York, UK}
  \country{}
}
\email{gricel.vazquez@york.ac.uk}

\author{Alexandros Evangelidis}
\orcid{0000-0003-4032-3042}
\affiliation{%
  \institution{University of York, UK}
  \country{}
}
\email{alexandros.evangelidis@york.ac.uk}

\author{Sepeedeh Shahbeigi}
\orcid{0000-0002-4405-5754}
\affiliation{%
  \institution{University of York, UK}
  \country{}
}
\email{sepeedeh.shahbeigi@york.ac.uk}

\author{Radu Calinescu}
\orcid{0000-0002-2678-9260}
\affiliation{%
  \institution{University of York, UK}
  \country{}
}
\email{radu.calinescu@york.ac.uk}

\author{Simos Gerasimou}
\orcid{0000-0002-2706-5272}
\affiliation{%
  \institution{\hspace{-5mm}Cyprus University of Technology, Cyprus}
  \country{}
}
\affiliation{%
  \institution{University of York, UK}
  \country{}  
}
\email{simos.gerasimou@cut.ac.cy    }

\renewcommand{\shortauthors}{V{\'a}zquez et al.}

\begin{abstract}

  Integrating Large Language Models (LLMs) into complex software systems enables the generation of human-understandable explanations of opaque AI processes, such as automated task planning. However, the quality and reliability of these explanations heavily depend on effective prompt engineering. The lack of a systematic understanding of how diverse stakeholder groups formulate and refine prompts hinders the development of tools that can automate this process. \change{CH3.1}{We introduce COMPASS (COgnitive Modelling for Prompt Automated SynthesiS), a proof-of-concept} self-adaptive approach that formalises prompt engineering as a cognitive and probabilistic decision-making process. 
  COMPASS models unobservable users' latent cognitive states, such as attention and comprehension, uncertainty, and observable interaction cues as a POMDP, whose synthesised policy enables adaptive generation of explanations and prompt refinements. We evaluate COMPASS using two diverse cyber-physical system case studies to assess the adaptive explanation generation and their qualities, both quantitatively and qualitatively. \change{CH3.2}{Our results demonstrate the feasibility of COMPASS integrating human cognition and user profile's feedback into automated prompt synthesis in complex task planning systems.}
\end{abstract}



\maketitle

\section{Introduction}
Large Language Models (LLMs) are increasingly integrated into software-intensive systems, providing new pathways for automation and effective collaboration between humans and AI~\cite{weyns2023towards}.
These pathways have the potential to revolutionise diverse industrial sectors, from reducing administrative burdens by automating medical documentation and subscription in healthcare~\cite{cascella2023evaluating} to streamlining decision-making processes by generating detailed reports and plans from raw data in finance~\cite{li2023large}, e-commerce~\cite{soviero2024chatgpt} and agriculture~\cite{sapkota2025multi}. 
In such complex data-intensive computational processes, however, generating human-understandable explanations that help users understand these otherwise opaque processes is equally important~\cite{jacovi-goldberg-2020-towards}.
These explanations are vital for ensuring transparency, trustworthiness, and effective human oversight, especially when the underlying AI components (used for task scheduling or probabilistic planning) are inherently opaque or difficult to interpret~\cite{li2020explanations}.

The quality, reliability and appropriateness of LLM-driven explanations critically rely on how prompts, i.e., the textual instructions guiding the LLM behaviour, are formulated and adapted to diverse user needs and contexts~\cite{10.5555/3692070.3692380}.
Despite growing interest in devising effective prompts, recent approaches rely on human intuition, ad hoc experimentation, or simplistic heuristics~\cite{li2024exploring}. 
The absence of a principled understanding of how different system stakeholders, e.g., AI experts, roboticists, project managers, and non-expert end users, iteratively construct, revise, and assess prompts hinders further LLM integration into adaptive systems. 

Since these systems are typically deployed in operational contexts involving diverse stakeholders, their informational needs, cognitive capacities, and interpretive preferences vary considerably~\cite{keil2006explanation}. 
Inevitably, these stakeholders require distinct explanation forms (varying in granularity, tone, and structure) to effectively comprehend the same information (e.g., a scheduling plan assigning tasks to a human-robot team). 
Also, user comprehension and attention dynamically vary, depending on cognitive factors such as workload, fatigue and domain familiarity~\cite{miller2019explanation}. 
For instance, explanations deemed appropriate in a previous interaction of a roboticist may become suboptimal or cognitively burdensome in another interaction occurring late during the shift of the same roboticist. 
Hence, static or generic prompts \chBlue{risk yielding} explanations that are overly technical, insufficiently informative, or misaligned with the user's cognitive state.
Addressing this challenge entails self-adaptation mechanisms that continuously align explanation generation with user feedback and cognitive variability~\cite{lyu2024towards,li2020explanations}.

\change{CH2}{While stakeholders could remain involved and explicitly articulate their explanation preferences, reliance on explicit preference elicitation places a non-trivial cognitive burden on users~\cite{kleinberg2024challenge}.
Also, well-documented cognitive biases, like asymmetric perception of losses and gains, and hyperbolic discounting of future rewards, often result in inconsistent and context-dependent preference elicitation~\cite{lyngs2018so}.  
Accordingly, a self-adaptive explanation system could mitigate these cognitive and status-quo biases~\cite{kahneman1991anomalies} by leveraging user profiles, prior knowledge of explanation preferences, and anticipated levels of understanding and attention to generate planner explanations without requiring additional input.
Although feedback from different stakeholders can be incorporated to refine and update this model, such feedback should remain intentionally optional~\cite{d2019learning}. 
This design choice reflects realistic deployment conditions, where users may be unwilling or unable to provide feedback (e.g., in call-centre settings, feedback requests are common, yet only a subset of users completes the corresponding feedback forms).}



We introduce \approach\ (COgnitive Modelling for Prompt Automated SynthesiS), a self-adaptive approach that formalises prompt engineering as a cognitive and probabilistic decision-making process. 
\approach\ integrates probabilistic reasoning and cognitive modelling to support the self-adaptive generation of task plan explanations using LLMs. More specifically, \approach\ leverages a Partially Observable Markov Decision Process (POMDP) to model the user's latent, probabilistic cognitive state, comprising unobservable variables such as attention and understanding, along with observable feedback from user interaction.
The POMDP analysis enables synthesising optimal prompt policies (comprising desired tone, level of detail, and structure) that maximise the expected acceptance and effectiveness of generated explanations by the target user.
\approach\ dynamically synthesises explanations, adapting their tone, structure, and granularity to the evolving cognitive profile and contextual user needs. 

Our \approach\ evaluation is based on two case studies in the construction and agriculture cyber-physical system (CPS) domains. 
We leverage both proprietary (OpenAI GPT-5, Google Gemini-2.5-Pro) and open-source (DeepSeek-V3.2) LLMs. 
\chBlue{Our results demonstrate} the feasibility of integrating LLMs to automatically generate planning problems, solutions, and explanations from natural language descriptions with minimal human input.
Further, through a user study with \change{A1.1}{32} participants, 
we \chBlue{extract initial insights} into human perception of different automatically generated explanations, and the effectiveness of adapting them to match user preferences and different cognitive states.
\chBlue{We consider these results as the basis for performing a more in-depth and longitudinal study on the effect of LLM-driven task plan explanations.}
We make the following concrete contributions:



\noindent$\bullet$ \change{CH3.3}{A conceptual} formalisation of prompt generation for task planners' explanations as a POMDP that captures both observable user interactions and unobservable cognitive states;
\\$\bullet$ The \approach\ self-adaptive explanation approach that dynamically adjusts LLM prompts and explanations based on user feedback, cognitive predictions, and contextual changes in task planning;
\\$\bullet$ A benchmark of 72 explanations tailored to three stakeholder profiles, with variations in tone, level of detail and format automatically generated state-of-the-art LLMs (OpenAI GPT-5, Google Gemini-2.5-
Pro, DeepSeek-V3.2); and
\\$\bullet$ The \approach\ \change{CH3.4}{proof-of-concept} validation in the CPS case studies of construction and agriculture, showing how adaptive explanations \change{CH3.5}{can influence} alignment, personalisation, and quality across diverse user profiles.

\section{Construction CPS Mission}
\label{sec:construction_case_study}
\textbf{Planning Problem}. We motivate \approach\ using a CPS construction scenario provided by our industrial partner. 
The scenario concerns task automation of a multi-level office building construction project, whose layout and tasks are obtained from a Building Information Model (BIM)~\cite{borrmann2018building}. 
The construction site comprises 10 distinct rooms (A-J) and involves interdependent tasks requiring coordinated execution between autonomous robots and a human worker. The tasks include:

\noindent $\bullet$
Foundation preparation - $t1$ ( rooms F and G): 
forming the initial structural base of the building, requiring heavy-duty work and precision, and assigned to robotic agents.

\noindent $\bullet$
Electrical installation - $t2$ (room H): 
laying out and connecting the building's electrical wiring and systems, requiring human expertise.

\noindent $\bullet$
Plumbing installation - $t3$ (rooms D and E): 
including the installation of pipes and fixtures for water distribution and drainage systems, assigned either to a specialised robot or the worker.


\noindent $\bullet$
Finishing work - $t4$ (rooms I and J): 
involving precision activities like painting and installing fixtures,  delegated to robots.

The CPS team comprises three robots ($r1, r2, r3$) and a worker ($h1$), each initially positioned at a distinct location on the construction site. Tasks can be retried a finite number in case of failure. Movement between any two connected locations incurs a uniform cost of 10 units.
Table~\ref{table:construction_costs_prob} shows task costs, success probabilities, durations, and allowed retries associated with each task. 
Costs are abstract units denoting resources consumed (e.g., energy, time).

Addressing this problem entails producing a plan that maximises the probability of successful mission completion while minimising operational costs.
We assume that an AI planner~\cite{pandey2016hybrid,valentini2020temporal} is employed to synthesise effective mission plans such that the minimum overall mission success probability is 95\%.



\begin{table}[]
\caption{Task cost, success probability, duration, and maximum number of retries per task and agent.}
\vspace{-4mm}
\small
\resizebox{0.47\textwidth}{!}{ 
\begin{tabular}{ll|llll}
\hline
\multicolumn{1}{c}{Agent (initial loc.)} & Tasks & Cost & Success prob. & Duration & Max. retries \\ \hline
\multicolumn{1}{c}{r1 (Room B)} & $[t1,t3]$ & [2,5] & [0.99,0.95] & [12,40] & [4,2] \\
\multicolumn{1}{c}{r2 (Room C)} & $[t1]$ & [2] & [0.98] & [13] & [4] \\
\multicolumn{1}{c}{r3 (Room J)} & $[t4]$ & [1] & [0.99] & [18] & [5] \\
\multicolumn{1}{c}{h1 (Room H)} & $[t2,t3]$ & [8,10] & [0.97,0.98] & [25,35] & [2,1] \\ \hline
\end{tabular}
}
\label{table:construction_costs_prob}
\vspace{-6mm}
\end{table}

\noindent
\textbf{Adaptive Explanation Problem}. 
The CPS mission involves heterogeneous stakeholder groups with distinct expertise and cognitive capabilities, categorised into the following representative profiles 
(i) AI planning and robotics experts;
(ii) domain experts (finance and administrative personnel);
and (iii) non-experts and end users. 
These groups differ significantly in their familiarity with planning models, risk interpretation, and technical understanding.
Hence, the investigated problem considers the adaptive generation of profile-tailored explanations for synthesised task plans.
Each explanation must consider the estimated likelihood of a user accepting or rejecting a given explanation, and predictions of the user’s cognitive state, i.e., their levels of attention and understanding. 


When a user requests a new explanation, feedback from prior interactions associated with that profile should be used to refine subsequent generations.
Accordingly, solving this problem entails continuously updating the explanation synthesis process as new feedback and state predictions become available, yielding a closed-loop process for adaptive and tailored explanation generation. 





\section{Background}
\label{sec:background}

\noindent
\textbf{LLM.} Large Language Models (LLMs), an application of generative AI (GenAI), are developed using probabilistic methods to generate human-like text based on extensive and varied datasets~\cite{radford2019language}. 
These models operate by estimating the likelihood of the next token in a sequence, given the previous context, an approach known as auto-regressive generation~\cite{vaswani2017attention, radford2018improving}. 
Widely used LLM-based tools include OpenAI's ChatGPT~\cite{openai2025chatgpt5,openai2024gpt4}, Google's Gemini~\cite{Google2025Gemini}, 
and DeepSeek~\cite{liu2024deepseek}. 
Prompting strategies for LLMs include zero-shot, one-shot and few-shots prompting, where none, a single or multiple examples are provided to guide the generated output~\cite{reynolds2021prompt}.

\vspace{1mm}
\noindent
\textbf{Markov decision process (MDP)}. A MDP is a stochastic model represented by the tuple \(\mathcal{M} = (S, s_I, A, P, L, R)\), comprising a finite set of states $S$; an initial state \(s_I \in S\); a finite set of actions \(A \neq \emptyset\); a partial probabilistic transition function \(P : S \times A \rightarrow \mathit{Dist}(S)\) that maps state-action pairs to discrete probability distributions over \(S\); a labelling function \(L: S \rightarrow 2^{\mathit{AP}}\) that maps each state to a set of atomic propositions $AP$; a reward function \(R\), defined as a tuple $R=(r_{state},r_{action})$ comprising a state reward function $r_{state}:S\rightarrow \mathbb{R}_{\geq 0}$ and an action reward function $r_{action}:S\times A\rightarrow \mathbb{R}_{\geq 0}$.

\vspace{1mm}
\noindent
\textbf{POMDP}. A POMDP is a tuple $M = (S, s_I,$ $A, P, L, R, O, \text{obs})$ where $(S, s_I, A, P, L, R)$ is an MDP; $O$ is a finite set of observations; $\text{obs} : S \to O$ is a labelling of states with observations; such that, for any states $s, s' \in S$ with $\text{obs}(s) = \text{obs}(s')$, their available actions must be identical, i.e., $A(s) = A(s')$. 
States in $S$ cannot be directly determined, only their corresponding observation $\text{obs}(s) \in O$~\cite{norman2017verification}. 
An observation-based strategy of a POMDP $M$
is a function \(\sigma : \text{FPaths}_M \to \text{Dist}(A)\) such that \(\sigma\) is a strategy of the MDP \((S, s_I, A, P, L, R)\); and for any paths \(\pi = s_0 \xrightarrow{a_0} s_1 \xrightarrow{a_1} \cdots s_n\), and \(\pi' = s_0 \xrightarrow{a_0'} s_1 \xrightarrow{a_1'} \cdots s_n\) satisfying \(\text{obs}(s_i) = \text{obs}(s_i')\) and \(a_i = a_i'\) for all \(i\), we have \(\sigma(\pi) = \sigma(\pi')\).

\vspace{1mm}
\noindent
\textbf{Probabilistic Computation Tree Logic  (PCTL)}. PCTL~\cite{hansson1994pctl,bianco1995model} is used to express quantitative properties over probabilistic models. State formulae are defined as:
$\Phi ::= \textit{true} \mid a \mid \neg\Phi \mid \Phi_1 \land \Phi_2 \mid \texttt{P}_{\bowtie p}[\varphi] \mid \texttt{R}^r_{\bowtie q}[\rho]$, 
where $a \in AP$ is an atomic proposition, $\bowtie \in \{<, \le, >, \ge\}$, $p \in [0,1]$, $q \ge 0$, and $q$ denotes a reward or cost structure. Path formulae are defined as: $\varphi ::= \mathsf{X}\Phi \mid \Phi_1\mathsf{U}^{\le k}\Phi_2 \mid \Phi_1\mathsf{U}\Phi_2$. 
Operators such as eventually ($\mathsf{F}\Phi$) can be derived from this ($ \textit{true}\mathsf{U} \Phi$). Query formulae such as ``what is the maximum/minimum probability of eventually reaching a state satisfying the predicate goal'' ($\texttt{P}_{\max/\min=?}[\mathsf{F}\,\mathsf{goal}]$), can also be defined for rewards $\texttt{R}_{\max/\min=?}$.

\begin{figure*}[t]
\centering
    \includegraphics[width=\textwidth]{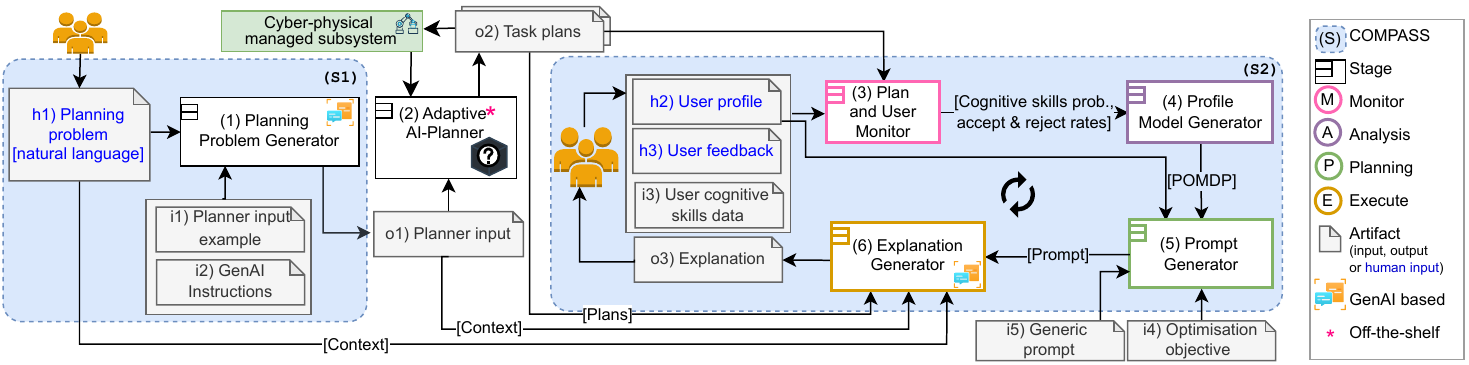}
  \vspace{-7mm}
  \caption{COMPASS stages and artefacts, showing the adaptation of explanations (S2) positioned within the MAPE-K loop.}
  \label{fig:overview}
  \vspace{-2mm}
\end{figure*}

\section{\approach}

An overview of \approach, which comprises six steps (1-6), is shown in Figure~\ref{fig:overview}. 
These steps are divided into two stages: \textbf{(S1)}~for the generation of task plans given the problem description in natural language, and \textbf{(S2)}~for the adaptive synthesis of human-readable profile-specific explanations of synthesised task plans. 
Table~\ref{tab:terminology} details the terminology used throughout to present \approach.

\begin{figure}[t]
    \centering
    \includegraphics[width=\linewidth]{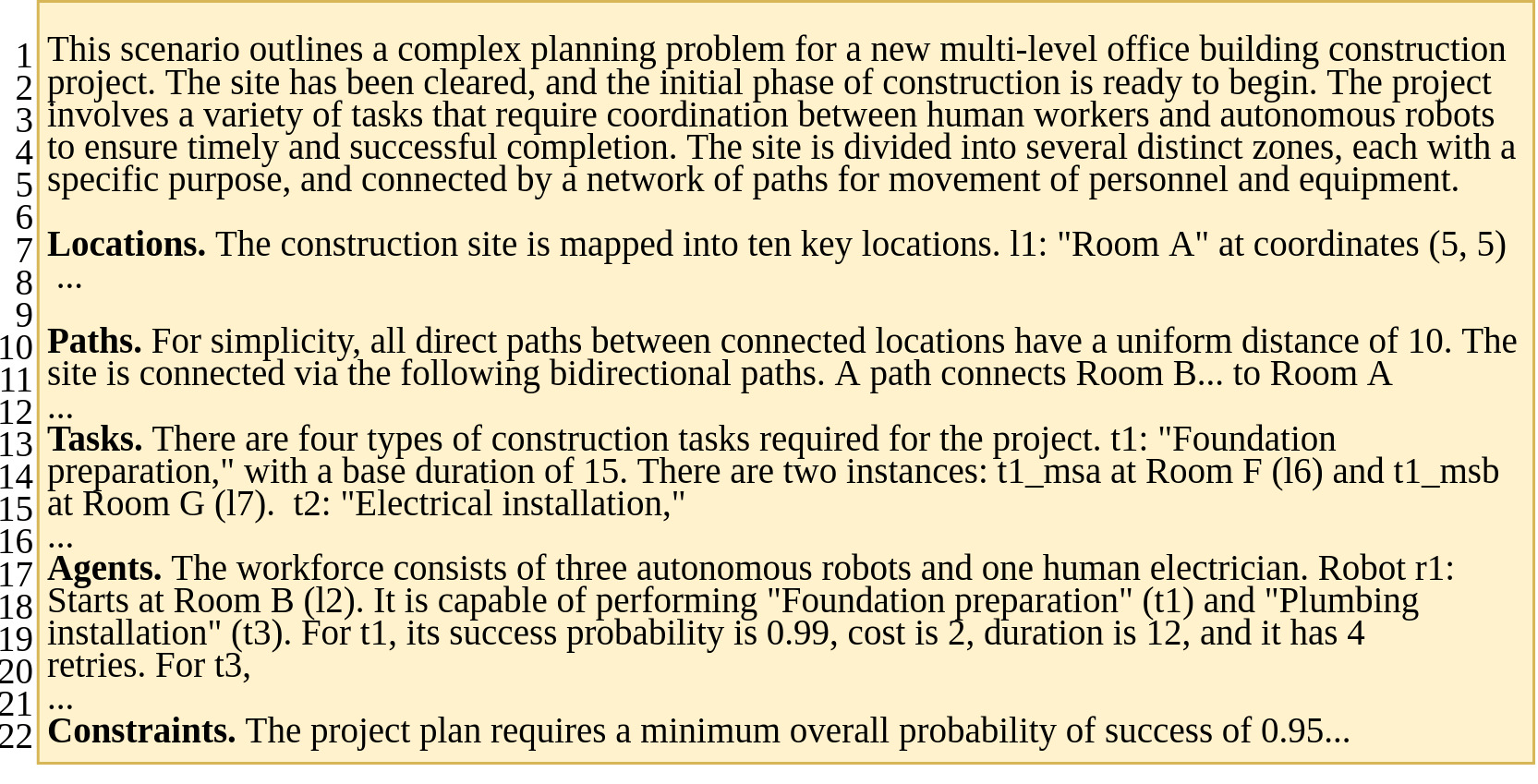}
    \vspace{-8mm}
    \caption{Planning problem description of scenario in Section~\ref{sec:construction_case_study} provided in natural language by stakeholders  ($h1$).}
    \label{fig:problem_natural_lang}
    \vspace{-3mm}
\end{figure}

\subsection{Problem Generation (S1)}
Stage (S1) begins by receiving a planning problem described in natural language, assembled from various stakeholders. 
\approach\ utilises GenAI and off-the-shelf AI planners~\cite{valentini2020temporal,vazquez2025adaptive} to convert this textual input into executable plans. 
First, our GenAI-based \textbf{Planning Problem Generator~(1)} receives three input files ($h1, i1, i2$), and generates the output of the selected planner ($o1$):

\noindent$\bullet$
Planning problem in natural language ($h1$): 
A description of the planning problem specified by stakeholders, including information about the available robots and human agents, tasks, locations, paths, and mission constraints. 
For instance, the description of the construction planning problem from Section~\ref{sec:construction_case_study} is shown in Figure~\ref{fig:problem_natural_lang}.
    
\noindent$\bullet$ One-Shot Planner input example ($i1$): 
A well-structured planning problem example that can be processed by the AI Planner, serving as a one-shot structural template~\cite{reynolds2021prompt} and defining the correct syntax and accepted format expected by the AI Planner. 
This input is mandatory since the expected format varies significantly across different planners. 
For instance, classical planners consume problem definitions written in the Planning Domain Definition Language (PDDL)~\cite{fox2003pddl2, younes2004ppddl1}, modern frameworks like the Unified Planning library accept Python-based problem specifications~\cite{unified_planning_UP}, while other planners may utilise custom DSLs and formats such as JSON~\cite{vazquez2025adaptive}. 
    
\begin{figure}[t]
    \vspace{-4mm}
    \centering
    \includegraphics[width=1\linewidth]{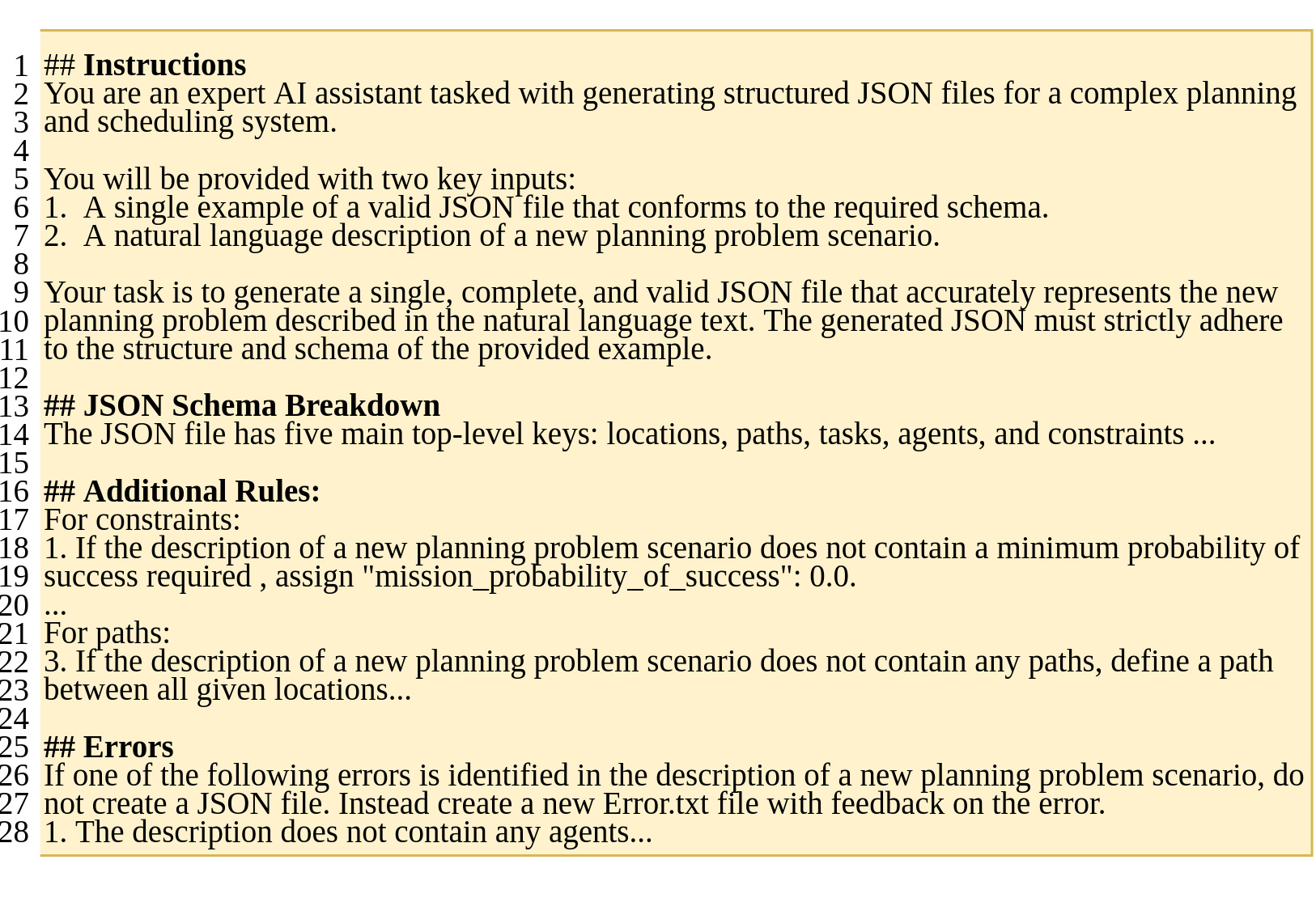}
    \vspace{-11mm}
    \caption{Example of GenAI instructions~($i2$).}
    \label{fig:prompt_input_i2}
    \vspace{-7mm}
\end{figure}

    
\noindent$\bullet$   GenAI instructions ($i2$): A crafted prompt with instructions for the Problem Generator to output a file aligned with the planner input example format $i1$. 
Figure~\ref{fig:prompt_input_i2} shows an example set of instructions to generate a JSON-formatted planner input.
These instructions detail the role of the GenAI component (lines 1-3), the supporting documents (lines 5-7), the overall goal (lines 9-11), and a breakdown of the different content parts of the expected output (lines 13-14).
Validity checks assess the completeness of $i1$ and compliance with the expected planner input (lines 17-27). 
For example, lines 18-19 instruct the execution of a typical validity check to ensure that the minimum required success probability is specified in $i1$.
Finally, error handling instructions are also  included (lines 26–28).

\begin{table}[t]
\caption{Terms and explanations used in \approach}
\vspace{-3mm}
\centering
\resizebox{0.49\textwidth}{!}{ 
\begin{tabular}{l p{0.48\textwidth}}
\hline
\textbf{Terminology} & \textbf{Definition} \\ \hline
user & An individual interacting with COMPASS \\ 
profile & A specific role or expertise of a user (e.g., researcher, roboticists, software developer, policymaker, educator, student). \\ 
profiled explanations & Explanations tailored to different users, e.g., generated using GenAI. \\ 
cognitive skills & Mental abilities, e.g., reasoning, understanding and attention. \\ 
cognitive skill level & Discrete level of a cognitive skill (e.g., low, medium, high).\\ 
cognitive state & The mental condition of the user, defined by the \textit{levels} of their cognitive skills (e.g., low attention and high understanding levels).\\
prompt & A piece of text or instruction given to a language model to elicit a specific response or generate content (see~\cite{vatsal2024survey}). \\ 
slots & A placeholder in a prompt that is completed through \textit{slot filling}~\cite{li2023generative}. \\ 
explanation & A GenAI generated summary that describes the reasoning behind the solution to a task planning problem in the context of COMPASS.\\
\hline

\end{tabular}
}
\label{tab:terminology}
\vspace{-4mm}
\end{table}

    
\noindent$\bullet$ Planner input ($o1$): The output from the Planning Problem Generator is a structured, machine-readable version of the natural language problem. 
The output can be represented as a function from tokens-to-tokens $o1 = f1(h1,i1,i2)$. 
Since this output is a direct product of a GenAI model, it is available for expert review to verify its correctness before further processing in \approach\ Stage 2.

Next, the \textbf{Adaptive-AI Planner~(2)} generates one or multiple feasible task plans~(o2) using a selected AI planner, represented as a function $o_2=f2(o_1)$. Example planners that can be utilised include fast downward~\cite{helmert2006fast}, NSPH~\cite{scala2016interval}, and Tamer~\cite{valentini2020temporal,micheli2021synthesis}. 
Within \approach, the AI planner is treated as a black-box, making our approach generic and agnostic to the selected planner. 
\change{B1}{Furthermore, the AI planner might internally have incremental mechanisms to generate new plans on-the-fly by adapting existing plans as changes or failures of the system are detected~\cite{pandey2016hybrid}. 
This is particularly useful when system changes are localised (e.g., in the plan minimum probability of success requirement); AI planners can leverage this aspect to incrementally and efficiently generate new plans by building on or adapting previously computed plans~\cite{dantam2016incremental,vazquez2025adaptive}.}


The successful generation of feasible task plans concludes stage S1. 
These plans are ready for deployment, and can be inspected by a human-in-the-loop or directly consumed and executed by the subject CPS~\cite{weyns2023self,camara2014stochastic}. 
The specific deployment mechanisms are beyond the scope of this work.

\subsection{Adaptive Synthesis of Plan Explanations (S2)}

During deployment in Stage (S2), the \textbf{Plan and User Monitor~(3)} \approach\ component monitors and responds to user requests for new explanations. 
The objective of this stage is to generate explanations ($o3$) from the available plans, considering the user's preferences. 
The stage starts by monitoring updates on the generated task plans, and changes to the following user data:

\noindent $\bullet$ User profile~($h2$): 
This encapsulates the characteristics of the end-user seeking an explanation (Table~\ref{tab:terminology}). 
Through profiling, \approach\ categorises individuals into distinct groups to enable modelling and adapting explanations based on each group's common expectations. 
For instance, robotics experts might prefer detailed insights into the sequence of tasks assigned to the deployed robots, whereas finance expers may prioritise explanations centred on plan-related costs.
    
\noindent $\bullet$ User feedback~($h3$): 
User feedback on preferred explanations is monitored and accumulated across profiles based on whether an explanation is accepted or rejected by a user. 
Prior knowledge based on past experience \change{B2}{is used to set the baseline for $h3$ (informing the initial initial POMDP model reward structure as explained later)}, while new feedback collected at runtime guides \approach\ in tailoring explanations for different profiles.

\noindent $\bullet$ User cognitive skills data ($i3$): 
In addition to user feedback $h3$, \approach\ assumes the availability of a cognitive skills predictor, likely through using a  Deep Learning model~\cite{gao2022deep}, that captures the probability that a user is in a specific cognitive level (e.g., high, medium or low attention level). 
\change{B3}{Following research like DeepDecs~\cite{calinescu2024controller,calinescu2022discrete}, these probabilities can be derived from the confusion matrix of DNN predictors.} 
This information is quite important and enable to combine user feedback (explicitly provided by the user themselves) with data predicted from DNNs trained on relevant data (capturing intangible and difficult to express information).

\change{A2}{Concerning the cognitive skills predictor, a substantial body of research in the ergonomics and human factors literature addresses the estimation of users’ cognitive states, such as attention and mental workload, from observable data~\cite{van2018understanding}. 
Prior work has explored how such cognitive variables may be operationalised and inferred using combinations of behavioural indicators and physiological signals~\cite{ayres2021validity}, task performance metrics, interaction logs, and non-invasive sensing modalities such as eye-tracking or heart-rate measures~\cite{dehais2020neuroergonomics}. Recent approaches leverage data-driven techniques to estimate cognitive states, with convolutional neural networks also being explored for this purpose~\cite{simfukwe2023cnn}. 
COMPASS leverages this past research and assumes the availability of such a predictor~\cite{gao2022deep}.}

\vspace{-2mm}
\subsubsection{Profile POMDP Model Generation}
\chBlue{
COMPASS frames prompt generation as a sequential decision-making problem under several uncertainty sources: (i) uncertainty in the user’s latent cognitive state (e.g., level of understanding, attention, or cognitive load); (ii) uncertainty due to imperfect predictors that estimate such cognitive states; and (iii) uncertainty in the user’s explanation acceptance, which may vary across users under the same observable profile.}
\change{CH1.1}{}

\chBlue{
A key modelling consideration is that a user’s internal cognitive condition cannot be directly observed or extracted explicitly~\cite{lyngs2018so}.
Instead, the system relies on noisy and partial observations derived from users' believed state, interactions and feedback. A POMDP naturally formalises this setting by: 
(i) representing unobservable cognitive variables as a hidden state; 
(ii)~maintaining a belief state, i.e., a probability distribution over possible cognitive states, rather than committing to a single estimate; and 
(ii)~modelling how prompt design choices (e.g., tone, level of detail, structure) are influenced by past acceptance data and the user’s underlying cognitive state. Simpler modelling alternatives, such as fully observable MDPs, rule-based systems, heuristic policies or  contextual bandits~\cite{langford2007epoch} entail direct access to users’ internal cognitive states or strong assumptions that the relationship between observations and actions is deterministic and well understood.}
\change{CH1.2}{}


To generate explanations, the \textbf{Profile Model Generator~(4)} constructs a profile-specific POMDP model representing the user’s cognitive skills and the likelihood of accepting or rejecting a generated explanation. 
In this model, the true cognitive skill level of the user are hidden variables; only the predicted by the DNN levels ($ie$) are known. 
The model further integrates user feedback on the acceptance rates based on previously presented explanations. 
The policy obtained by solving the POMDP denotes the prompt elements that inform the LLM to produce an explanation appropriate to the user's profile and its anticipated cognitive skills level.

In the POMDP, the user's cognitive skill levels are not directly observable. 
Instead, a \textit{belief} state is maintained and the presumed levels are derived from an imperfect DNN. 
We define a cognitive state as a collection of $k$, $1\leq k \leq K$ skills, where $cogSkill_k$ represents the $k$-th real cognitive skill and $cogSkillPred_k$ its predicted level, both classified into $M$ discrete levels. 
For example, if $cogSkill_1$ is attention and $M=3$, we can classify a user's attention level into low, medium or high.  
\textbf{Listing~\ref{lst:observables}} defines these predictions as observable variables in the PRISM language. A support variable $step$ is used to control the execution flow of the POMDP.



 The \emph{Turn} module, shown in \textbf{Listing~\ref{lst:controlFlow}}, controls the model evolution by allowing only one action at a time. Let $P$ be the number of prompt elements (slots) to fill in a prompt by solving the POMDP. 

\noindent$\bullet$
$S\_total=K+P+1$ is the number of transitions in the POMDP that control the flow of the model. The $+1$ adds an end state to ensure the inclusion of the final transition reward (self-transitions of terminal sink states are excluded).

\noindent$\bullet$ \textit{done} is a Boolean formula that denotes the termination point of the model flow, occurring when ``the prompt has been filled''.

The Turn module synchronises with other modules in the POMDP and manages the flow of actions using the counter variable $step$. 
Synchronisation occurs through transitions sharing the same action label, denoted within square brackets `[]', e.g., ``ObserveCogSkill\_1''. 
The first $K$ transitions set the $K$ cognitive skill levels of the user; the following $P$ transitions correspond to informing the policy synthesis to select prompt elements.

\begin{lstlisting}[float, frame=tb, belowskip=-1.8\baselineskip, language={Prism}, numbers=left, rulesepcolor=\color{black}, rulecolor=\color{black}, breaklines=true, breakatwhitespace=true, firstnumber=1, firstline=1, linewidth=\hsize, %xleftmargin=9pt, 
        caption={Observable state variables.}, 
        label={lst:observables}, 
    aboveskip=0pt,
    belowskip=-2mm,
    xleftmargin=4pt,
    xrightmargin=0pt
            ]
pomdp
observables
    step, cogSkillPred_1, ..., cogSkillPred_K
endobservables
\end{lstlisting}

\begin{lstlisting}[float, frame=tb, belowskip=-1.8\baselineskip, language={Prism}, numbers=left, rulesepcolor=\color{black}, rulecolor=\color{black}, breaklines=true, breakatwhitespace=true, firstnumber=1, firstline=1, linewidth=\hsize, %xleftmargin=9pt, 
        caption={Turn module controls the actions flow.}, 
        label={lst:controlFlow}, 
    aboveskip=0pt,
    belowskip=-2mm,
    xleftmargin=4pt,
    xrightmargin=0pt
            ]
const int S_total = K + P + 1; // total number of transitions
formula done = step = S_total; // end of model flow
module Turn
    step: [0..S_total] init 0;
    [ObserveCogSkill_1] step=0 -> 1:(step'=1);
    ...
    [ObserveCogSkill_K] step=k-1 -> 1:(step'=k);
    [SelectPrompt_1_1]    step=k -> 1:(step'=k+1);
    ...
    [SelectPrompt_p_q]    step=k+p-1 -> 1:(step'=k+p);
    [End]               step=S_total-1 -> 1:(step'=S_total);
endmodule
\end{lstlisting}

\begin{figure}[t]
\centering
\begin{lstlisting}[
    language=Prism,
    numbers=left,
    frame=tb,
    breaklines=true,
    breakatwhitespace=true,
    firstnumber=1,
    firstline=1,
    rulesepcolor=\color{black},
    rulecolor=\color{black},
    caption={Cognitive skills prediction module.},
    label={lst:humanCognitive-model},
    aboveskip=0pt,
    belowskip=0pt,
    xleftmargin=4pt,
    xrightmargin=0pt
]
const int profile_1 = 1;
...
const int profile_N = N;
const int profile;          // current user profile
const int level_1 = 1;     // cognitive level (low)
...
const int level_M = M;     // cognitive level (high)
// For profile 1, Cognitive skill 1:
const double P_1_1_L1L1, P_1_1_L1L2, ..., P_1_1_LMLM;
...
// For profile N, Cognitive skill K:
const double P_n_K_L1L1, P_n_K_L1L2, ..., P_n_K_LMLM;
module CognitiveSkillLevelPrediction
    cogSkill_1: [1..M];      // true, hidden cognitive skill
    cogSkillPred_1: [1..M];  // observable prediction
    ...
    cogSkill_K: [1..M];
    cogSkillPred_K: [1..M];
    // Probabilistic skill transitions based on profile
    // For each profile n from 1 to N, and each cognitive skill k from 1 to K:
    [ObservecogSkill_k] profile=profile_n ->
        P_n_k_L1L1:(cogSkill_k'=level_1)&(cogSkillPred_k'=level_1) +
        P_n_k_L1L2:(cogSkill_k'=level_1)&(cogSkillPred_k'=level_2) +
        ...
        P_n_k_LMLM:(cogSkill_k'=level_M)&(cogSkillPred_k'=level_M);
endmodule
\end{lstlisting}%
\vspace{-5mm}
\end{figure}

The \emph{Cognitive skill level prediction} module (\textbf{Listing~\ref{lst:humanCognitive-model}}) selects the user's unobservable and predicted cognitive skill levels. 
Given user profile $n$ (lines 1-3), cognitive skill $k$ with a value of $M$ possible levels (lines 5-7), this module defines the joint probability $P\_n\_k\_LiLj$ 
\[
P_{n,k(i,j)}=P(cogSkill_k=level_i,\ cogSkillPred_k=level_j\ |\ \ profile_n)
\]


These values  (lines 8-12) represent the probability that the user truly has cognitive skill $k$ at level $i$, while the predicted level for the same skill is $j$ (lines 8-12). 
The Cognitive Skills Prediction module consists of state variables for each cognitive skill level ($cogSkill_k\leq M$) and its prediction ($cogSkillPred_k\leq M$)~(lines~14-18).
Transitions in this model are used to probabilistically set the user's skill levels and predicted level (lines 19-25). 
Only transitions matching the user profile are available, as specified by the guard $profile=profile\_n$ (line~21). 
Lines 22~25 show the selection of one out of the $M$ possible cognitive skill levels (for both the true and predicted values) of cognitive skill $k$, for a user with profile $n$.

By building and analysing the POMDP model, \approach\ finds an optimal policy that, given a user profile and the user's skill level, maximises the acceptance likelihood of an explanation generated from a prompt. 
An action in such a policy represents the selected value of a prompt element (i.e., a prompt slot). 

The \emph{``acceptance'' reward structure} (\textbf{Listing~\ref{lst:PRISM-rewards}}) defines the utility gained when a user accepts a specific prompt option, conditional on their profile and cognitive skill levels.
Let $U=cogSkill_1,...,cogSkill_K$ be the vector representing the user's $K$ cognitive skill levels. 
The model assigns a reward based on the function $Utility(n,p,q,U)$, representing the utility for a user with profile $n$ accepting option $q$ for prompt slot $p$, given their true cognitive skill levels are represented by the vector $U$ (with values from 1 to $M$). 

The reward ``acceptance" structure assigns a utility value when a transition is taken.
Multiple transitions might exist with the same label $SelectPrompt\_p\_q$, shown in lines 5 and 9. The assignment is guarded by the user's profile and their specific cognitive skills combination. 
Since the latter are hidden variables, these transitions are indistinguishable from the model's analysis perspective. 
Hence, the model cannot assign a single, definite reward for the observed action. 
Instead, each potential utility is weighted by the current probability in the model's belief state that the user possesses the corresponding vector of hidden cognitive skills. 
The complete module contains one such rule for every possible combination of user profiles, prompt selections, and cognitive skill vectors.


\textbf{Utility function.} 
\approach\ adopts Prospect Theory principles~\cite{kahneman2013prospect, tversky1992utility} to define the utility function \change{CH4.1}{for supporting the cognitive model.}
The utility of an accepted prompt is not a fixed value; it is a subjective gain evaluated relative to a specific reference point.
Concretely, this reference point represents a baseline utility $b_{min}$ (Listing~\ref{lst:formulas-rewards} line 3). 
We initialise this value to 5 because we want to capture the fact that a suboptimal explanation for a user is better than receiving no explanation at all. 
We also set a value of $20$ for the maximum utility that can be achieved ($b_{max}$, (Listing~\ref{lst:formulas-rewards} line 4)).
We assign these values because we want to have a range of utility values
sufficient enough that will allow to distinguish between the various policies effectively. Note that, from a mathematical point of view, changing $b_{min}$ and $b_{max}$ will not have any effect on the computed policy.
To construct the reward structure in PRISM, we employ Bayesian learning~\cite{zhao2024bayesian}, leveraging the beta-binomial model to estimate user acceptance probabilities from limited data.
First, we define $r_{n,p,q} \in [0,1]$ which represents the probability that a user with
profile $n$ accepts an explanation generated using option $q$ for prompt $p$. 
We then use the estimated probability $r_{n,p,q}$ in the computation of the utility using the prospect-theoretic formulas defined in lines 18-22 of Listing~\ref{lst:formulas-rewards}. The general form is:
$
    utility = b_{\min} + \kappa \cdot (b_{\max} - b_{\min})\cdot r_{n,p,q}^{\alpha},
$
%
%
where $\alpha$ denotes a parameter based
on the concept of diminishing returns from Prospect Theory~\cite{tversky1992utility}. This parameter controls how sensitive the utility of a prompt is with respect to changes in its acceptance probability. We set $\alpha = 0.88$ following the guidelines from~\cite{tversky1992utility}.
Intuitively, a value of $0.88$ ensures that improvements to poorly-performing prompts are valued higher than equivalent improvements to already high-performing ones. For example, if a prompt's acceptance rate increases from $20\%$ to $30\%$, the utility gain will be higher compared to a prompt that increases from $95\%$ to $98\%$.

In Listing~\ref{lst:formulas-rewards}, we define three utility functions that capture different 
degrees of alignment between the chosen prompt setting and the user's latent cognitive state:

\noindent$\bullet$
    $utility\_n\_p\_q\_match$: when the chosen prompt setting aligns with the latent state of the user

    \noindent$\bullet$ $utility\_n\_p\_q\_okay$: when the chosen prompt setting partially aligns with the latent state of the user

    \noindent$\bullet$ $utility\_n\_p\_q\_mismatch$: when there is a mismatch between the prompt setting and the latent state of the user

The role of $\kappa$ is to penalise cases where there is a misalignment between the prompt we provide to the user and the user's latent cognitive state. 
For example, a $\kappa=1.0$ indicates a perfect match, and as a result, we do not apply a penalty, i.e., the original calculated utility is retained.
We assign values of 0.75 and 0.5 to $\kappa$ to penalise partial and total mismatches, respectively.
In the former case, $75\%$ of the potential gain is retained and $50\%$ in the latter.

\textcolor{black}{The acceptance and rejection rates are obtained from user feedback. Although using binary accept/reject feedback provides only partial information about user reactions, this choice
reflects realistic deployment scenarios in which detailed or continuous feedback is often unavailable or burdensome to collect.}



\begin{lstlisting}[float, frame=tb, belowskip=-1.8\baselineskip, language={Prism}, numbers=left, rulesepcolor=\color{black}, rulecolor=\color{black}, breaklines=true, breakatwhitespace=true, firstnumber=1, firstline=1, linewidth=\hsize, %xleftmargin=9pt, 
        caption={Reward structure for the expected acceptance}, 
        label={lst:PRISM-rewards},
            aboveskip=0pt,
    belowskip=-2mm,
    xleftmargin=4pt,
    xrightmargin=0pt
            ]
rewards "acceptance"
    ...
    // For each profile n,
    // for each slot p with option q, 'Slot_p_q' and
    [SelectPrompt_p_q] (profile=profile_n) & 
    (cogSkill_1 = L1) & ... & (cogSkill_N = L1):
                utility_n_p_q_u1;
    ...
    [SelectPrompt_p_q] (profile=profile_n) & 
    (cogSkill_1 = LM) & ... & (cogSkill_N = LM):
                utility_n_p_q_ux;
endrewards
\end{lstlisting}

\begin{lstlisting}[float, frame=tb, belowskip=-1.8\baselineskip, language={Prism}, numbers=left, rulesepcolor=\color{black}, rulecolor=\color{black}, breaklines=true, breakatwhitespace=true, firstnumber=1, firstline=1, linewidth=\hsize, %xleftmargin=9pt, 
        caption={Utility formulas for rewards, and their constants}, 
        label={lst:formulas-rewards},
            aboveskip=0pt,
    belowskip=-4mm,
    xleftmargin=4pt,
    xrightmargin=0pt
            ]
const double alpha0 = 1;
const double beta0 = 1;
const double b_min = 5;
const double b_max = 20;
const double kappa_okay = 0.75;
const double kappa_mismatch = 0.5;
// For each user profile n
const double alpha = 0.88;
...
//For each prompt slot p, option q
const int r_accepted_p_q;
const int r_rejected_p_q;
...
// For each prompt slot p, for user n
formula r_n_p_q = (r_accepted_p_q + alpha0) / 
    (r_accepted_p_q + r_rejected_p_q + alpha0 + beta0);

formula utility_n_p_q_match = b_min + (b_max - b_min) * pow(r_n_p_q, alpha);

formula utility_n_p_q_okay = b_min + kappa_okay * (b_max - b_min) * pow(r_n_p_q, alpha);

formula utility_n_p_q_mismatch = b_min + kappa_mismatch * (b_max - b_min) * pow(r_n_p_q, alpha);

\end{lstlisting}

\vspace{0mm}
\subsubsection{Prompt and Explanations Generation}
\label{sec:Prompt and Explanations Generation}

\begin{figure}[t]
    \vspace{-0.5mm}
    \centering
    \includegraphics[width=\linewidth]{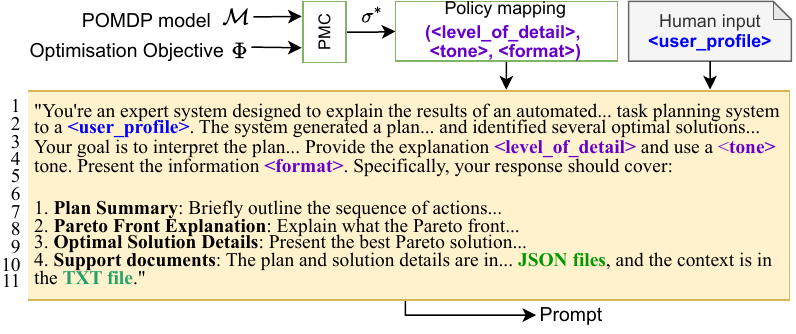}
    \vspace{-7mm}
    \caption{Prompt generator~(5) flow. See also Table~\ref{tab:actions_2_prompt_mapping}.}
    \label{fig:PromptGenerator}
    \vspace{-4mm}
\end{figure}

The generation of a tailored explanation is initiated by the \textbf{Prompt Generator (5)}, which formally synthesises an optimal policy $\sigma^*$ from the POMDP model $M$ using probabilistic model checking (Figure~\ref{fig:PromptGenerator})~\cite{calinescu2018efficient,gerasimou2018synthesis}. 
Given the POMDP $\mathcal{M}$ and a property maximising the expected acceptance of an explanation ($\phi=R$\{``acceptance"\}$_{max=?}$ [F done]), the optimal policy satisfying $\phi$ is computed. 
The resulting policy maps belief states to actions, providing the optimal choice for each slot in our prompt template. 
Internally, the \textbf{Prompt Generator} also contains a \textit{Prompt Selection Module} that constructs the final prompt by populating the slots with the choices dictated by $\sigma^*$. 

The instantiated prompt serves as a comprehensive set of instructions for the LLM, and includes directives to process and integrate data from various sources: the problem context and planner input ($h1$ and $o1$, respectively), the most recent plan generated by the planner ($o2$), and the user's profile. 
Figure~\ref{fig:PromptGenerator} shows an example of such a prompt, demonstrating how the actions selected in the POMDP correspond to parameters such as the level of detail, tone, and format of the prompt
(whose values are given in Table~\ref{tab:actions_2_prompt_mapping}).

\begin{table}[t]
\caption{Mapping from POMDP actions to prompt options and user profiles for the construction case study. IDs show the POMDP variable and real-valued instantiation.}
\label{tab:actions_2_prompt_mapping}
\vspace{-4mm}
\centering
\small
\resizebox{0.48\textwidth}{!}{ 
\begin{tabular}{lp{8cm}}
\hline
\textbf{ID} & \textbf{Text in prompt for \textcolor{blue}{<user\_profile>}} \\ \hline 
profile1 & AI expert (assumes deep understanding of optimisation, probability...) \\ 
profile2 & Domain expert (in finance or administration) (assumes focus should be on cost, and high-level risk) \\ 
profile3 & Non-expert (assumes minimal technical knowledge, requiring analogies and simple terms) \\ \hline

\textbf{ID} & \textbf{Text in prompt for \textcolor{electricviolet}{<level\_of\_detail>}} \\ \hline
p1\_q1 & in high detail (long answer) \\ 
p1\_q2 & in a concise summary with only the necessary details to understand the main points (short answer) \\ \hline

\textbf{ID} & \textbf{Text in prompt for \textcolor{electricviolet}{<tone>}} \\  \hline
p2\_q1 & precise, technical, formal and precise tone \\ 
p2\_q2 & casual, conversational and simple tone (use examples if needed) \\ \hline

\textbf{ID} & \textbf{Text in prompt for \textcolor{electricviolet}{<format>}} \\  \hline
p3\_q1 & in a step-by-step list \\ 
p3\_q2 & as a single coherent paragraph, as a summary, no bullet points nor list \\ 
p3\_q3 & as a series of bullet points highlighting key items (avoid paragraphs) \\ \hline
\end{tabular}
}
\vspace{-2mm}
\end{table}

By combining the dynamically selected slot choices with this contextual data, the prompt guides the \textbf{Explanation Generator (6)} to produce a relevant and personalised explanation ($o3$) tailored to the user's profile, inferred cognitive state, and the current task-planning scenario.
Given this information, the \textbf{Explanation Generator} leverages GenAI to produce a tailored explanation aligned with the selected policy, ensuring that the generated content considers the user’s cognitive state, their profile preferences, and \textit{contextual} factors. Context information about the planning problem and results is obtained from the initial planning problem~($h1$), the planner's input file~($o1$), and the generated plan(s)~($o2$).

\vspace{2mm}
\textbf{Example 1}. Assume that the policy computed by \approach\ selects the following actions for the decision points in the prompt: $\langle p1\_q1, p2\_q1, p3\_p3 \rangle$ (see reward actions in Listing~\ref{lst:PRISM-rewards}). According to the mapping in Table~\ref{tab:actions_2_prompt_mapping}, this yields an instantiated prompt with: 
\{level of detail: high level, tone: formal, format: bullet point\}.


\textcolor{black}{We use POMDPs due to the several uncertainties shaping a user's likelihood of accepting an explanation. 
The joint POMDP structure maintains a belief over the latent user states and selects parameter values that maximise the expected acceptance probability.}


\subsubsection{Adaptive synthesis}

\approach\ adapts its explanations in response to new information, ensuring that they remain aligned with the user's preferences and (predicted and real) cognitive state. 
The adaptation mechanism is triggered by the Plan and User Monitor~(3), and involves updating the POMDP model and re-synthesising the optimal policy. The primary triggers for adaptation are:
%

\noindent $\bullet$ 
    Change in user feedback ($h3$): When a user accepts or rejects an explanation, the system updates the corresponding $R_{acceptance}$ and $R_{rejection}$ counts for that user's profile (Listing~\ref{lst:formulas-rewards}). This alters the POMDP reward structure (Listing~\ref{lst:PRISM-rewards}), potentially leading to a new optimal policy that better reflects the updated user preferences.
    
\noindent $\bullet$ 
    Change in cognitive state prediction ($i3$): As the system gathers more data on the user's interactions, the underlying cognitive skills predictor may provide updated probabilities for the user's cognitive state ($i3$).  
    These new probabilities update the transition dynamics in the POMDP (Listing~\ref{lst:humanCognitive-model}),  refining the model's belief about the user's hidden state and prompting a re-evaluation of the optimal policy.

\noindent $\bullet$ Change in context ($h2$): 
An adaptation is also triggered if a new user with a \emph{different profile} requests an explanation \change{CH6.1}{(Listing~\ref{lst:humanCognitive-model}, line 4)}, if the Adaptive AI Planner generates \textbf{a new or modified plan} ($o2$) in response to changes in the operating environment, or when a brand \emph{new planning problem} is presented ($h1$). This ensures the generated explanation is always relevant to the current user and the latest version of the task plan.

\textbf{Example 2}. Assume that a non-expert user requests an explanation, resulting in a policy that maps to: \{\textit{level of detail}: summarised, \textit{tone}: casual, \textit{format}: list\}. 
Later, an AI expert \chBlue{uses COMPASS and} requests an explanation \change{CH6.2}{triggering a change in context (i.e., change in profile $h2$). COMPASS modifies the POMDP model setting the parameter value $profile$ (see Listing~\ref{lst:humanCognitive-model}, line 4) to a new value representing the AI expert profile. After verifying the updated POMDP, the new policy results in prompt:} \{\textit{level of detail}: high-detailed, \textit{tone}: technical, \textit{format}: paragraph style\}. 
\chBlue{After some time, a new change is detected when the ``AI expert profile" receives feedback as multiple rejections for the generated explanation which inform \approach\ to update the rejection rates (line 11 in Listing~\ref{lst:formulas-rewards}).} 
This results in a different explanation: \{level of detail: high-detailed, tone: technical, format: step-by-step list\}. 
Finally, at the end of a long working day, the AI expert has a predicted lower level of attention; this changes the probabilities on the prediction of the user's cognitive state (lines 12 in Listing~\ref{lst:humanCognitive-model}). 
A new explanation is hence generated with prompt: {level of detail: low (or concise), tone: technical, format: bullet points}. 

\subsection{Implementation}
We instantiate \approach\ using the adaptive task planner from~\cite{vazquez2025adaptive}. The planner example~($i1$) and generated problem file ($o1$) are JSON-formatted; instructions~($i1$) are given as a single text file (Figure~\ref{fig:prompt_input_i2}). 
The planner generates a JSON file of multiple task plans~($o2$) showing Pareto-optimal trade-offs between cost and success probability (see details in~\cite{vazquez2025adaptive}). 
For the Prompt Generator~(5), we use the PRISM model checker to extract the POMDP policy. 
Our \approach\  tool and evaluation results are available at our~\href{https://github.com/Gricel-lee/COMPASS-LLM}{GitHub} website~\change{A3}{\cite{GitHub}}.

\section{Evaluation}
\label{sec:evaluation}

\vspace{-1mm}
We evaluate COMPASS through two distinct multi-robot multi-human CPS case studies: a construction scenario introduced in Section~\ref{sec:construction_case_study}, and an agricultural domain scenario adapted from~\cite{vazquez2025adaptive}.


\vspace{-3mm}
\subsection{Research Questions}

\textbf{RQ1 (Problem Generation Feasibility): 
Can the planning problem generator create correct planner inputs?} 
We examine whether the Planning Problem Generator (Stage S1) produces both syntactically compliant (with the required JSON format) and semantically valid outputs, yielding solvable planning problems.

\noindent
\textbf{RQ2 (Adaptation): How effective is \approach\ in dealing with monitored changes by adapting the generated explanation?} 
We investigate \approach's end-to-end ability to dynamically reconfigure explanations based on monitored changes in user interactions, shifts in cognitive state and acceptance rates.

\noindent
\textbf{RQ3 (Alignment): How well do the generated and adapted explanations align with the initial prompt and the user's provided feedback, respectively?} 
Since plan explanations are tailored to individual users' needs, we conduct a user study with \change{A1.2}{32} participants to evaluate \chBlue{if \approach} can yield explanations that adhere to explicit constraints such as tone, format, and detail. 

\noindent
\textbf{RQ4 (Personalisation): Does the Explanation Generator of \approach\ produce explanations tailored to individual stakeholder profiles?}
We conduct a second user study to assess whether the intention of the generated explanation to be informative for a specific user profile truly matches the survey participants' opinions.

\vspace{-2mm}
\subsection{Experimental Setup}

\textbf{RQ1 (Feasibility)}. We generated 20 variants per case study (randomly selecting 3-5 tasks, 1-3 robots, 1-3 humans, and 5-9 locations). For each variant, a planner input (JSON) file is created using different LLMs (Google Gemini-2.5-Pro~\cite{Google2025Gemini}, OpenAI GPT-5~\cite{openai2025chatgpt5}, and DeepSeek-V3.2~\cite{liu2024deepseek}). 
We assess feasibility by checking whether the JSON is correctly formatted (parsable), whether the AI planner can successfully interpret the data and extract the necessary information (processable), and whether plans are generated (feasible).

\noindent
\textbf{RQ2 (Adaptation)}. We present a \chBlue{proof-of-concept} adaptation scenario for the construction scenario, where all changes are present, as done in related research~\cite{gerasimou2014efficient}. We generated plans from the original case study, initially for a \textit{non-expert profile}. The initial explanation is then adapted based on changes in the user profile, explanation acceptance rates, and predictions of the user's cognitive state.

\noindent
\textbf{RQ3 (Alignment)}. We generated explanations for all combinations of profiles (three), two levels of detail, two tone types, and three explanation formats, yielding 36 explanations (per case study) using Google Gemini-2.5-Pro. 
We used a Google Forms survey to collect responses to predefined questions (\textit{Q}).  
First, each participant was shown an explanation obtained from a prompt with the following options: (i)~non-experts: summary, casual, list formatted; (ii)~AI experts: high detail, formal, paragraph formatted, and (iii)~domain experts (finance): high detail, formal, bullet points formatted. These were derived from the optimal POMDP policy, initialised with probabilities and rewards informed by domain knowledge.

Since we know the prompt from which the initial explanation was generated (ground truth), we assess how participants \textit{perceived} the explanation with the following questions: 
\textit{(Q1) How would you describe the level of detail?}; 
\textit{(Q2) How would you rate its tone?}; and 
\textit{(Q3) How would you describe its format?}, with nominal-categorical answers outlined in Table~\ref{tab:actions_2_prompt_mapping}. 
We generated confusion matrices based on these results. 
Participants were also asked to rate, on a scale from 1 to 5 (5 is better), the appropriateness of the \textit{level of detail}~(Q4), \textit{tone}~(Q5), and \textit{format}~(Q6) of the explanation. 
These ratings were later compared with their evaluations on the adapted explanation.

Next, participants were invited to select their \textit{preferred level of detail}, \textit{tone}, and \textit{format}. 
A second personalised explanation was then generated based on this feedback. 
As before, a set of questions was shown with ordinal numerical answers from 1 to 5 (5 is better): 
\textit{(Q7) Is the tone better?};
\textit{(Q8) Is the level of detail better?}; \textit{(Q9) Is the format better?}; and 
\textit{(Q10) Did the new explanation help you better understand the described solution?} 
Finally, participants were asked to provide demographic information.

\noindent
\textbf{RQ4 (Personalisation)}. 
A second survey classified all explanations according to the user profile (AI expert, domain expert, or non-expert) that participants believed they were intended for. 
All explanations from RQ3 were used in the survey (with all combinations of user profiles, level of detail, tone, and format) and shown to participants in a randomised order. 
Participants were then asked to select the user profile most closely matching the subject of the explanation (or the \textit{unsure} option). 
The goal of this survey was to evaluate whether the intended personalisation cues embedded in the LLM-driven explanations were perceived as such by independent evaluators. 
This enabled assessing the effectiveness and clarity of the personalisation strategies used, as well as identifying potential mismatches between the target and perceived audiences per explanation. 
For each explanation, we calculated the percentage of correct matches to the intended user profiles and then computed an overall average that is reported in the results. 
We also analysed the correctly classified explanations for each user profile to gain further insights into which profiles were easier to identify.

\subsection{Results and Discussion}
\textbf{RQ1 (Feasibility).} All evaluated LLMs (Gemini, GPT-5, DeepSeek) successfully generated parsable and processable JSON files for all 20 variants in both construction and agricultural case studies. 
This outcome demonstrates that \approach's Planning Problem Generator, when shown a one-shot example, can reliably produce syntactically valid and semantically-interpretable input structures compatible with the \approach-instantiated Adaptive AI Planner~\cite{vazquez2025adaptive}.

A detailed inspection of the generated JSON files confirmed that all relevant information from the original problem descriptions was accurately preserved; 
for instance, the number of robots and humans, their respective capabilities, and the bidirectional paths. 
The LLMs also correctly instantiated the task entries, i.e., task copies derived from other general tasks but localised at specific coordinates. 
This result shows full alignment between the generated data structures and the LLMs’ interpretive `understanding' of the presented problems, with a plausible explanation being that LLMs have been extensively trained to produce structured JSON outputs~\cite{mao2025prompts}.


From these generated JSON inputs, the AI planner successfully created feasible plans for all agricultural variants, while difference emerges in the construction case study. 
DeepSeek-V3.2 consistently produced feasible plans across all 20 variants, whereas GPT-5 and Gemini-2.5-Pro succeeded in only 4 out of 20 cases each. 
Further analysis revealed that the issue was not due to LLM limitations but to incomplete problem definitions. 
For example, in a problem variant with 3 tasks, 1 human, 1 robot, and 6 locations, none of the agents could perform the ``construction finishing work''.

These results signify that while LLMs can \change{CH3.6}{often} generate syntactically correct planner inputs $o1$ (JSON), the AI planner may fail to produce plans if the problem definition is incomplete. 
\approach\ users must thus ensure that problem descriptions are fully specified, e.g., via a human-on-the-loop check between stages S1 and S2. 
\change{B7.1}{Using other prompting strategies beyond one-shot could potentially alleviate the issues affecting GPT-5 and Gemini.}

\begin{figure}[t]
    \centering
    \includegraphics[width=\linewidth]{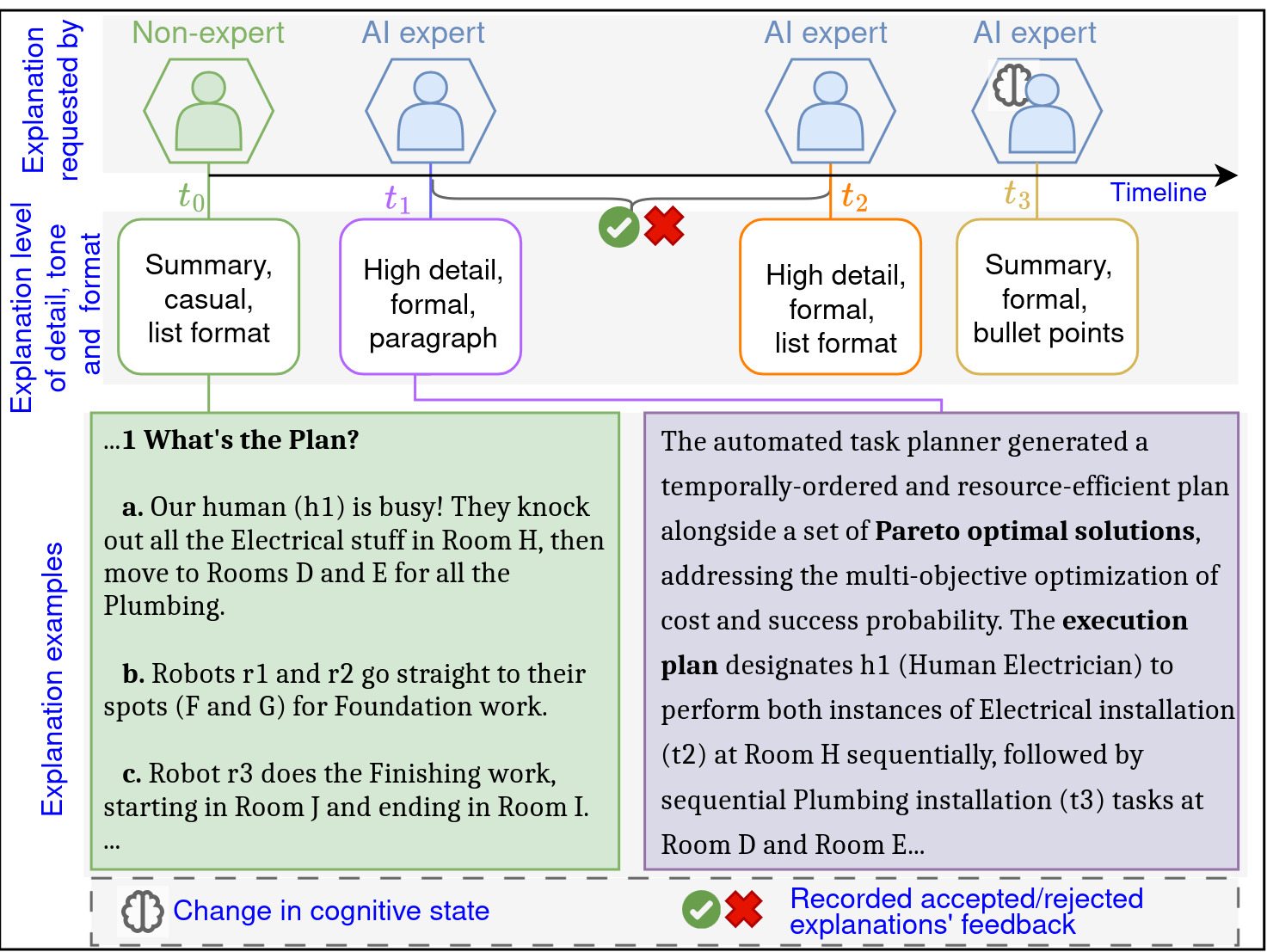}
    \vspace{-6mm}
    \caption{Scenario showing adaptive explanations}
    \label{fig:adaptation}
    \vspace{-5mm}
\end{figure}

\vspace{1mm}\noindent
\textbf{RQ2 (Adaptation)}. We show the adaptation process for a concrete scenario of the construction case study, assuming a continuous adaptation loop for the changes monitored by \approach: change in user profile, acceptance rates and predicted cognitive state of user (Figure~\ref{fig:adaptation}).
The first explanation at time  $t_0$ is generated when a non-expert user ($profile=3$, Table~\ref{tab:actions_2_prompt_mapping}) logs into the construction logistics system and requests an explanation on the currently available plans from the AI planner. 
\approach\ modifies the POMDP by assigning the corresponding profile value (line 4, Listing~\ref{lst:humanCognitive-model}). 
The new POMDP policy obtained, $\langle p1\_q2, p2\_q2, p3\_q1 \rangle$, yields a prompt with instructions to generate a \textit{summarised, casual tone} and \textit{list formatted} explanation (see Table~\ref{tab:actions_2_prompt_mapping}).

At time $t_1$, a new user (AI expert) logs in ($profile=1$), reflected in the corresponding profile value change (line 4, Listing~\ref{lst:humanCognitive-model}).  
The AI expert requests an explanation, based on which \approach\ generates a new policy $\langle p1\_q1, p2\_q1, p3\_q2 \rangle$, which in turns produces a \textit{high-detailed, technical} and as a \textit{paragraph style} explanation.

At time $t_2$, the same AI expert (from $t_1$) logs in and requests a new explanation. 
During this period, the system has collected feedback from other AI expert users who also requested explanations. 
\approach\ updates the acceptance number from 29 to 34, and rejections from 8 to 57, for explanations generated from option $p3\_q2$ (lines 11-12, Listing~\ref{lst:formulas-rewards}). 
Also, \approach\ yields a new POMDP policy $\langle p1\_q1, p2\_q1, p3\_q1 \rangle$, changing the format from paragraph style to \textit{step-by-step} list. 
This adaptation shows how \approach\ employs user feedback to iteratively refine explanations; here, optimising the explanation format to better suit expert user preferences.

Next, the same user logs in after a long working day (time $t3$). The cognitive state predictor used to estimate the user's cognitive skills ($i3$ in Figure~\ref{fig:overview}) estimates a higher probability that the user's attention is lower due to the time of the day. 
\approach\ updates the POMDP transition probabilities (line 12, Listing~\ref{lst:humanCognitive-model}), resulting in a new policy $\langle p1\_q2, p2\_q1, p3\_q3 \rangle$, changing the level of detail from \textit{high} to \textit{low}, and the format from list to \textit{bullet points}. 
Such modifications aim to \textcolor{black}{minimise interaction effort, and thus mitigate potential} cognitive fatigue, by reducing the volume of information shown and emphasising key points in a visually analysable structure. 
By simplifying the content and highlighting essential information, COMPASS helps sustain the user’s attention and comprehension even under reduced cognitive capacity, ensuring that explanations remain accessible and actionable in low-attention contexts.

This representative scenario illustrates how COMPASS \change{CH3.7}{could support the} runtime adaptation of explanation generation to maintain user profile preferences and user's attention engagement.  
\change{CH4.2}{Although COMPASS does not directly measure cognitive load via self-report instruments, its design explicitly targets the reduction of interaction-related cognitive effort. 
This design choice is grounded in Prospect Theory, which characterises human decision-making as reference-dependent and sensitive to cognitive effort, losses, and diminishing returns. By proactively adapting explanation granularity and structure to inferred attention and understanding levels, COMPASS aims to avoid explanations that users would view as cognitively costly relative to their current reference point.
}


\vspace{1mm}\noindent
\textbf{RQ3 (Alignment)}. 
We analyse the survey results using descriptive statistics and quantitative data analysis as per~\cite{weyns2023self}. 

\vspace{0.2mm}
\noindent
\textbf{Population and sampling}. Our target population are individuals from academia and industry classified into three profiles.
We analysed anonymised data from: 15 AI experts, 4 domain experts (in finance and management), and 13 non-experts in these areas.

\vspace{0.1mm}
\noindent
\textbf{Survey results}. 
Table~\ref{tab:rq3_constr_confision_matrices} presents the findings from questions $Q1$-$Q3$, comparing the real prompt attributes (tone, level of detail, and format) against those perceived by participants for the construction case study. 
Most participants correctly identified the tone, but discrepancies emerged in the detail levels (``summary" and ``detailed"). Evidently, without a direct reference point for what constitutes a ``short" or ``long" explanation, participants found it difficult to evaluate the level of detail. Regarding format, the paragraph style was consistently well recognised, but list- and bullet-formatted explanations exhibited greater overlap. After reviewing the generated explanations, we found that some list outputs included numbering (e.g., ``1, 2, 3" or ``a, b, c") alongside bullet symbols ($\cdot$), which may have introduced perceptual noise and contributed to this confusion. These \change{CH3.8}{findings indicate} the need to refine prompt construction to reduce ambiguity in perceived attributes. Specifically, prompt templates should more clearly distinguish levels of detail and enforce consistent formatting rules, ensuring that automatically generated explanations align more closely with users’ expectations across tone, detail, and format, and any other choices.

Tables~\ref{tab:rq3_const}a-f summarise this data from participants’ ratings on the initial and adapted explanations collected from $Q4$-$Q9$. 
For example, two non-experts ranked the level of detail as very appropriate (selecting 5, from 1-5), for their initial explanation, increasing to 4 non-experts for the final adapted explanation. 
A shift toward greater acceptance after adaptation is visualised for non-experts and AI experts, as the number of yellow and white cells shifts to the left post-adaptation; however, the opposite happens for domain experts. 
This is better visualised in Tables~\ref{tab:rq3_const}g-i, showing the number of individual participants that increased, decreased or did not change their score post-adaptation. 
For non-experts and AI experts, the total number of increased scores (16 and 21, respectively) is higher compared to no-change (10 and 14) or decreased scores (13 and 10). 
This shows that the adaptive explanation was perceived as better, although no consensus was reached among participants. 
However, this trend is reversed for non-experts, where a decrease in score was registered 7 times, compared to 3 increases and 2 equal scores. 
Further studies with a larger participant pool with this profile are needed to confirm these trends. 
Similar trends are observed for the agricultural case study (Tables~\ref{tab:rq3_agri_confision_matrices} and \ref{tab:rq3_agri}). 

Overall, these \change{CH3.9}{exploratory results yield indicative evidence that \approach\ can influence} user satisfaction and perceived appropriateness among non-experts and AI experts. \change{CH3.10}{The observed trends across user groups and case studies suggest that tailoring explanations to users' expertise could improve acceptance in AI planners, though these effects should be interpreted cautiously.}

These findings should \chBlue{also} be interpreted probabilistically, consistent with the probabilistic nature of our underlying POMDP model. 
For all profiles, \approach\ can incorporate these acceptance and rejection scores to modify the POMDP rewards accordingly. 
Hence, after this adaptation occurs, both non-experts and AI experts are \textit{more likely} to accept explanations similar to the adapted one whereas \chBlue{domain} experts are not.

\begin{figure}[t]
    \centering
    \includegraphics[width=\linewidth]{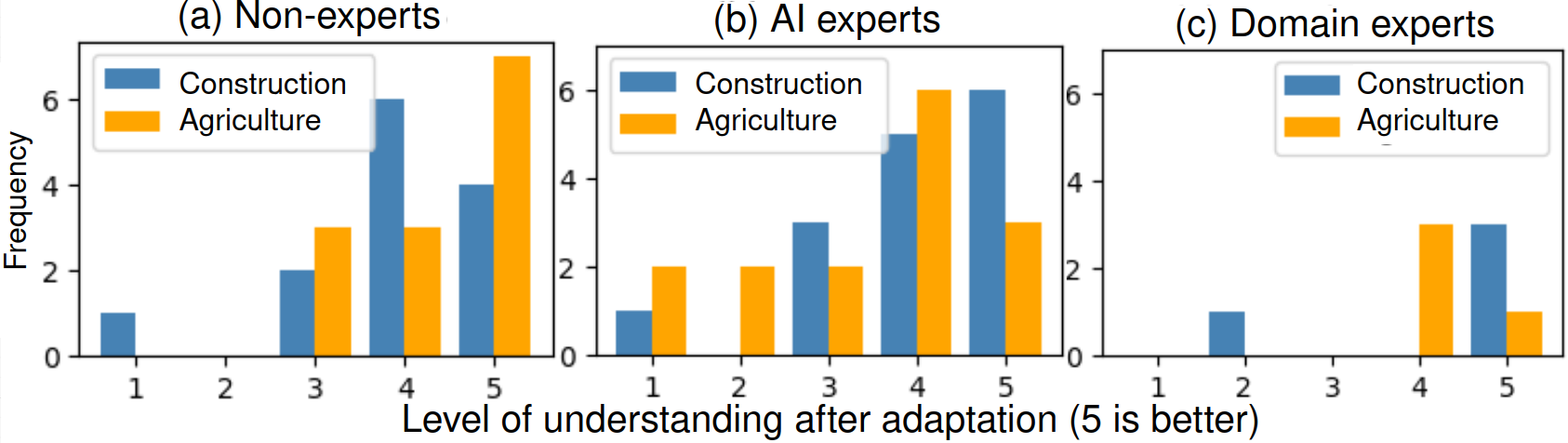}
    \vspace{-8mm} 
    \caption{Understanding level for adapted explanation}
    \label{tab:rq3_changeInUnderstanding} 
    \vspace{-6mm}
\end{figure}

For $Q10$, we obtained scores on how the new explanation helped participants better understand the planner's solution. 
Figure~\ref{tab:rq3_changeInUnderstanding} shows a consistent trend towards increasing understanding level for the adapted explanation compared to the initial. 
Both construction and agriculture tasks showed similar distributions; most participants reported moderate to high increase (scores 3–5). 
For domain experts, even when previous results show a decrease in the acceptance score (especially in the explanation format), almost all participants in this group provided a 4/5 score.
Overall, these \change{C3.11} {results suggest that} adaptive explanations \chBlue{could} support users' understanding from different expertise levels and domains.

\begin{table}[t]
\caption{Confusion matrices for ground truth (column) vs perceived (row---P) explanations for the construction study}
\vspace{-4mm}
\centering

\scriptsize
\begin{tabular}{lll|lll|llll}
\toprule
\multicolumn{3}{c}{\textbf{Tone}} &
\multicolumn{3}{c}{\textbf{Level of Details}} &
\multicolumn{4}{c}{\textbf{Format}}\\
\midrule
&$\!\!\!$\textbf{Cas.} &$\!\!\!$\textbf{Form.} 
    & &$\!\!\!$\textbf{Summ.} &$\!\!\!$\textbf{Det.}
    & &$\!\!\!$\textbf{Par.} &$\!\!\!$\textbf{List} &$\!\!\!$\textbf{Bul.}\\
$\!\!\!$\textbf{Cas. (P)} & 11   & 4
    &$\!\!\!$\textbf{Summ. (P)} & 12 & 12 
    &$\!\!\!$\textbf{Par. (P)} & 8 & 2 & 1\\
$\!\!\!$\textbf{Form. (P)} & 2 & 15
    &$\!\!\!$\textbf{Det. (P)} & 1 & 7
    &$\!\!\!$\textbf{List (P)} & 4 & 4 &1\\
&&&&&&$\!\!\!$\textbf{Bull. (P)} & 3 & 7 & 2\\
\bottomrule
\multicolumn{10}{l}{
Cas: Casual; Form: Formal; Summ: Summary; Det: Detailed; Par: Paragraph; Bull: Bullets}
\end{tabular}
\label{tab:rq3_constr_confision_matrices}
\vspace{-4mm}
\end{table}

\begin{table}[t]
    \centering
    \caption{Construction case study acceptance score}
    \vspace{-4mm}
    \label{tab:rq3_const}
    \includegraphics[width=\linewidth]{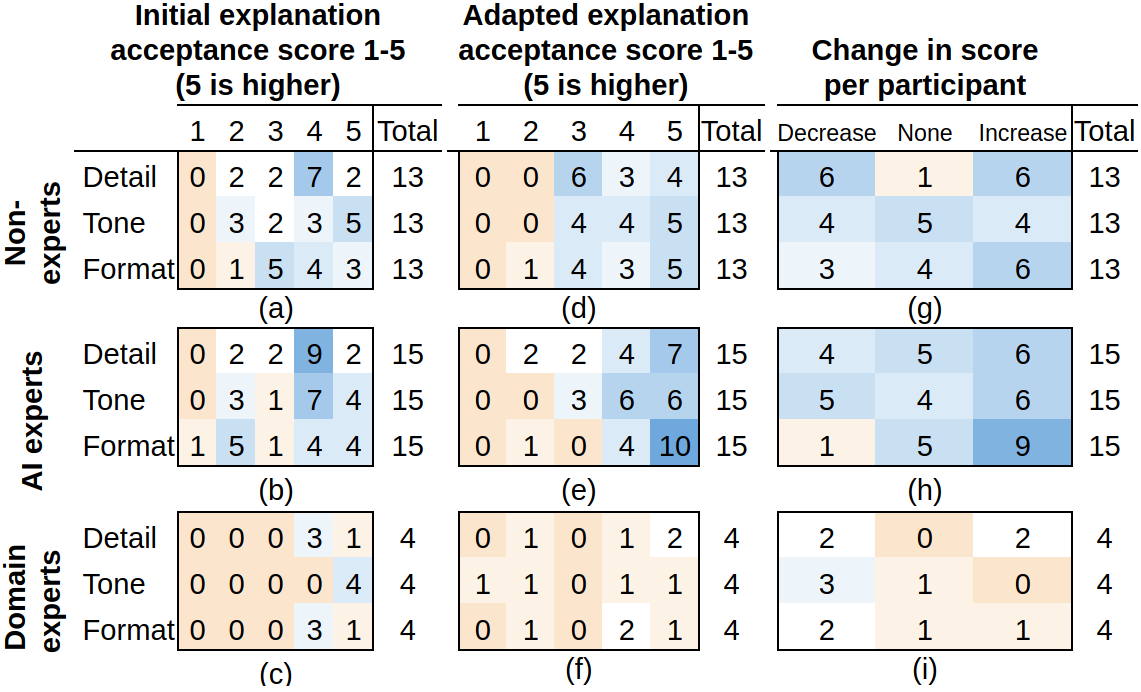}
    \vspace{-10mm}
\end{table}

\vspace{1mm}\noindent
\textbf{RQ4 (Personalisation).} Similar to RQ3, we used descriptive statistics and quantitative data analysis to analyse these survey results.

\noindent
\textbf{Population and sampling}. We analysed anonymised data from five participants with a computer science background working in academia (excluding participants from RQ3). 
Each participant classified all 72 explanations gathered from both case studies.

\begin{table}[t]
\caption{Confusion matrices for ground truth (column) vs perceived (row---P) explanations for the agriculture study}
\vspace{-4mm}
\centering

\scriptsize
\begin{tabular}{lll|lll|llll}
\toprule
\multicolumn{3}{c}{\textbf{Tone}} &
\multicolumn{3}{c}{\textbf{Level of Details}} &
\multicolumn{4}{c}{\textbf{Format}}\\
\midrule
&$\!\!\!$\textbf{Cas.} &$\!\!\!$\textbf{Form.} 
    & &$\!\!\!$\textbf{Summ.} &$\!\!\!$\textbf{Det.}
    & &$\!\!\!$\textbf{Par.} &$\!\!\!$\textbf{List} &$\!\!\!$\textbf{Bul.}\\
$\!\!\!$\textbf{Cas. (P)} & 8   & 5
    &$\!\!\!$\textbf{Summ. (P)} & 9 & 6 
    &$\!\!\!$\textbf{Par. (P)} & 8 & 3 & 0\\
$\!\!\!$\textbf{Form. (P)} & 5 & 14
    &$\!\!\!$\textbf{Det. (P)} & 4 & 13
    &$\!\!\!$\textbf{List (P)} & 5 & 3 &2\\
&&&&&&$\!\!\!$\textbf{Bull. (P)} & 2 & 7 & 2\\
\bottomrule
\multicolumn{10}{l}{
Cas: Casual; Form: Formal; Summ: Summary; Det: Detailed; Par: Paragraph; Bull: Bullets}
\end{tabular}
\label{tab:rq3_agri_confision_matrices}
\vspace{-3mm}
\end{table}

\begin{table}[t]
    \centering
    \caption{Agricultural case study acceptance scores}
    \vspace{-4mm}
    \label{tab:rq3_agri}
    \includegraphics[width=\linewidth]{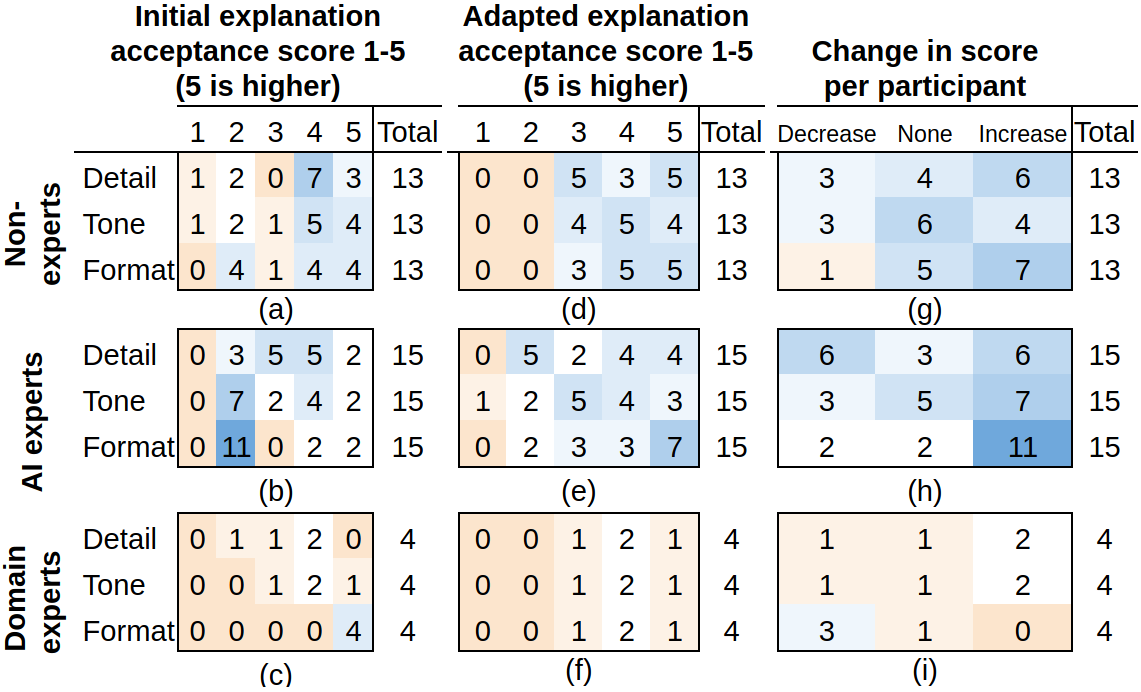}
    \vspace{-9mm}
\end{table}
%

\noindent
\textbf{Survey results.} For the construction case study, 49.8\% (35.6\% std) responses correctly matched the user profile for which the explanation was generated. 
The value increased for the agricultural case study (62.4\% with 40.1\% std).
As there are three possible profile options from which participants could select (AI expert, domain expert, non-expert),
a random selection would result in 33.33\% correct matches. 
Thus, on average, participants correctly identified the intended user profile at a rate substantially above chance. 
These results suggest that some profile personalisation cues embedded in the explanations were generally perceivable by participants, though variability was high, as indicated by the standard deviations.


To provide further insights into the misclassification per profile, we computed the proportion of explanations correctly classified for each user profile, and the proportion of explanations for which participants selected the \textit{unsure} option. In the construction study, domain expert explanations were the easiest to identify (55.0\% correct), followed by non-expert explanations (36.7\%) and AI experts (44.0\%). 
The unsure option (11.1\%) signified a light ambiguity level.

In contrast, the agricultural case study showed higher correct classification rates for AI experts (90\% accuracy), domain experts (56.7\%) and the unsure option~(11.7\%); with the lowest percentage also reported for non-experts (23.3\%). 
Overall, these results demonstrate that participants found it easier to recognise AI expert explanations in both settings, while non-expert explanations were consistently more difficult to classify across both domains.

A plausible explanation for this insight is that the generated plans from the AI planner are defined in technical terms such as ``Pareto optimality", ``probability of mission success", ``task-completion expected cost", and ``task identifiers", which might be perceived as tailored for AI experts. 
Another explanation is that we did not have participants with finance domain expertise to assess explanations intended for this profile. Thus, the accuracy of this profile detection may be underestimated. The survey results show that personalisation of AI explanations is not always detectable by independent evaluators, particularly when the content remains highly technical.

\section{Threats to Validity}

\textbf{Internal validity} threats concern causal inferences within our study. 
The effectiveness of \approach\ in refining prompts may vary with different LLMs (e.g., with stronger reasoning or instruction-following abilities) and different prompting strategies.
An advanced LLM might produce strong explanations even from weak prompts, while a light LLM could fail even with a perfect prompt, thus confounding \approach's measured impact. 
We mitigated this threat by generating AI planner input files using three state-of-the-art LLMs to compare performance across multiple planning problems in two case studies. 
The prompts in Figures~\ref{fig:prompt_input_i2} and~\ref{fig:PromptGenerator} were iteratively refined. 
We also acknowledge that LLMs are prone to hallucinations and output variability, even under identical prompts. 
Further, the binary accept/reject feedback in the POMDP provides only a coarse signal of user response, reflecting realistic operational settings where fine-grained or continuous feedback is rarely available or costly to obtain.
Finally, we evaluate the feasibility of the PDDL file generation step as an initial validation of correctness.
Verifying the extraction process, specifically, if information from the human input $h1$ is preserved, would strengthen our results and is thus part of our future work as in~\cite{vyas2025extensive, tantakoun2025llms}.


To mitigate \textbf{external validity} threats affecting the generalisation of \approach, our evaluation employed two CPS domains. 
Further, we validated both case studies through collaboration with industrial partners, ensuring practical relevance and applicability. 

\textbf{Conclusion validity} threats might arise given the relatively small sample size, reducing statistical inference. 
While trends in acceptance and perceived level of detail, tone and format were observed, results should be interpreted as indicative rather than conclusive. 
Replication with larger participant pools, especially more domain experts, would strengthen our findings' robustness.


\section{Related Work}
\label{sec:rel-work}
\textbf{LLMs for explainability and planning}. 
LLMS have been used to generate explanations for black-box models~\cite{kroeger2023llms, bhattacharjee2024towards} and to provide interactive explanations through conversational systems \cite{slack2023talktomodel, shen2023convxai}.
In medical contexts, LLMs support diagnostic reasoning~\cite{attai2024enhancing} and generate interpretable explanations to help the overall clinical decision-making ~\cite{umerenkov2023deciphering}.
In finance, LLMs generate explanations that improve the transparency of credit scoring models~\cite{feng2023empowering} and risk assessment~\cite{luz2024enhancing}.
Approaches like Tree-of-Thoughts~\cite{yao2023tree}, ``inner monologue''~\cite{huang2023innermonologue} and LLM+P~\cite{liu2023llmp} use LLMs as planning engines. 
Our work is complementary: rather than using an LLM to compute plans, COMPASS models human behaviour to adapt explanations of plans produced by formal planning methods.
\\
\textbf{LLMs for explainable AI (XAI).}
Within XAI, LLMs are powerful tools that
can translate complex and often cryptic AI model behaviours into easy-to-understand explanations~\cite{bilal2025llms,zhao2023explainability}.
LLMs have also been used to convert numeric outputs from traditional XAI methods, such as SHAP~\cite{lundberg2017shap} and LIME~\cite{ribeiro2016lime}, into natural language explanations~\cite{zytek2024explingo, burton2023nle}.
For further details, see the survey in~\cite{bilal2025llms}.
\\
\change{C1}{
\textbf{Explainable AI Planning (XAIP)}. XAIP is 
concerned with generating explanations for automated planning~\cite{chakraborti2020emerging}. 
Foundational work on model reconciliation formalises explanation as resolving mismatches between an agent’s planning model and a human’s mental model~\cite{chakraborti2019plan}, enabling planners to answer contrastive queries~\cite{krarup2019model,fox2017explainable,cashmore2019towards}. 
Expectation-aware planning further integrates human expectations into plan synthesis, often trading optimality for interpretability~\cite{sreedharan2020expectation}. In contrast, COMPASS adaptively tailors LLM-generated plan explanations by modelling users’ cognitive states and feedback, focusing on presentation-level adaptation across user profiles rather than modifying the planning process itself.
}
\\
\textbf{Explainability for self-adaptive systems (SAS).}
Explainability in SAS poses unique challenges due to their architecture, combining the managed and managing systems linked via feedback loops~\cite{straub2025explainabilitysas}. 
Existing work~\cite{bencomo2012sas,parra2021explainabilitysas} uses runtime models to provide explanations for the adaptation decisions. 
Closer to \approach,  \cite{alharbi2021explainability,li2020explanations} employ probabilistic modelling to personalise how explanations are delivered to the user.  
Unlike \approach, which synthesises explanations dynamically considering hidden cognitive states, they assume that personality traits are observable and explanations are static.
%
\\
\textbf{Modelling human behaviour via POMDPs.} 
POMDPs have been used to model decision-making under uncertainty by constructing models of human behaviour~\cite{calinescu2026verification}. 
Illustrative examples include: 
PsychSim~\cite{pynadath2005psych} for modelling complex social interactions and cognitive states, 
an assistance robot in nursing homes to infer the users' hidden goals~\cite{pineau2003pomdp}, and 
a POMDP for hand-washing aid for individuals with dementia~\cite{hoey2013pomdps}.
Contextual bandits~\cite{langford2007epoch} offer an alternative by modelling uncertainty in action outcomes conditioned on observable context~\cite{li2010contextual}.
However, they assume that context fully captures the user state and model single-step decisions~\cite{langford2007epoch}
In contrast, COMPASS uses POMDPs to model latent cognitive states and encode how actions influence future hidden states and rewards.

\vspace{-4mm}
\section{Conclusions and Future Work}

\approach\ is a proof-of-concept self-adaptive approach for generating task-plan explanations via LLMs. By treating prompt engineering as a cognitive and iterative process, \approach\ integrates user feedback, cognitive-state prediction, and context to tailor explanations to diverse users. 
We evaluated \approach\  in two industrial CPS, showing that LLMs can translate natural language problems into valid task-planning inputs and that adaptive explanations improve user understanding. 
Future work involves 
refining prompt instructions using zero- and few-shot prompting, supporting additional properties~\cite{menghi2022mission,vazquez2024robotics}, and incorporating additional planners~\cite{valentini2020temporal,gerasimou2021evolutionary}. 

\vspace{0.5mm}
\noindent
\textbf{Acknowledgements.} This research was supported by the EU Horizon projects AI4Work/SOPRANO (101135990/101120990), the ARIA-funded project ULTIMATE, and the Centre for Assuring Autonomy.

\bibliographystyle{ACM-Reference-Format}
\bibliography{bibfile}


\end{document}